# SCALING OPEN-WEIGHT LARGE LANGUAGE MODELS FOR HYDROPOWER REGULATORY INFORMATION EXTRACTION: A SYSTEMATIC ANALYSIS

**Hong-Jun Yoon**
Oak Ridge National Laboratory
Oak Ridge, TN 37830
yoonh@ornl.gov

**Faisal Ashraf**
Oak Ridge National Laboratory
Oak Ridge, TN 37830

**Thomas A. Ruggles**
Oak Ridge National Laboratory
Oak Ridge, TN 37830

**Debjani Singh**
Oak Ridge National Laboratory
Oak Ridge, TN 37830

November 14, 2025

## ABSTRACT

Information extraction from regulatory documents using large language models presents critical trade-offs between performance and computational resources. We evaluated seven open-weight models (0.6B-70B parameters) on hydropower licensing documentation to provide empirical deployment guidance.

Our analysis identified a pronounced 14B parameter threshold where validation methods transition from ineffective (F1 < 0.15) to viable (F1 = 0.64). Consumer-deployable models achieve 64% F1 through appropriate validation, while smaller models plateau at 51%. Large-scale models approach 77% F1 but require enterprise infrastructure.

We identified systematic hallucination patterns where perfect recall indicates extraction failure rather than success in smaller models. Our findings establish the first comprehensive resource-performance mapping for open-weight information extraction in regulatory contexts, enabling evidence-based model selection.

These results provide immediate value for hydropower compliance while contributing insights into parameter scaling effects that generalize across information extraction tasks.

## 1 Introduction

Large language models (LLMs) have transformed information extraction methodologies from unstructured textual data; however, a significant methodological paradox emerges when these systems exhibit ostensibly optimal performance metrics: specifically, instances where high recall values paradoxically indicate systematic error propagation rather than extraction accuracy. This statistical anomaly—wherein elevated recall coefficients correlate with hallucination phenomena rather than extraction precision—constitutes a fundamental methodological limitation affecting LLM deployment in critical regulatory compliance frameworks and validation protocols.

This extraction paradox demonstrates heightened significance within domain-specific technical documentation, where information extraction errors potentially precipitate substantial regulatory non-compliance and environmental impact assessment inaccuracies. Hydropower environmental mitigation licensing documentation presents an archetypal case study: these documents represent critical components within renewable energy infrastructure development frameworks, yet their inherently complex, semi-structured organizational schema renders precise information extraction simultaneously essential for regulatory compliance and methodologically challenging from a natural language processing perspective.

Hydropower represents a significant component of renewable and sustainable energy portfolios in the United States, contributing approximately one-third of all renewable energy to the US electric grid and constituting more than 6% of total electricity generation nationwide Uria Martinez and Johnson [2023]. This resource plays a crucial role in maintaining equilibrium between energy security imperatives and environmental conservation objectives. Notwithstanding its advantages, hydropower generation exhibits inherent associations with substantial ecological impacts, thereby necessitating rigorous regulatory frameworks and mitigation protocols. The Federal Energy Regulatory Commission (FERC) exercises jurisdictional authority over the licensing of non-federal hydropower facilities, issuing licenses with typical durations of 30–50 years. These regulatory instruments incorporate comprehensive environmental mitigation requirements designed to conserve biodiversity, enhance water quality parameters, facilitate fish passage and migration, improve public outdoor recreational opportunities, and maintain ecosystem structure and function.

Historically, FERC's environmental regulatory framework evolved significantly following the implementation of major environmental legislation, including the National Environmental Policy Act (NEPA) of 1970, the Clean Water Act (CWA) of 1972, and the Endangered Species Act (ESA) of 1973. A critical regulatory transformation occurred with the enactment of the Electric Consumers Protection Act (ECPA) of 1986, which mandated that FERC give equal consideration to power generation and environmental protection factors during licensing procedures. Subsequently, environmental mitigation requirements have expanded substantially in both complexity and scope, with further regulatory challenges emerging from the increased jurisdictional involvement of multiple state and federal regulatory bodies.

While these mitigation requirements are essential for ecological sustainability, the manual extraction of pertinent information from licensing documents presents significant operational challenges. Licensing documents are characterized by their unstructured nature and are typically composed in complex, domain-specific terminology without standardized nomenclature. Consequently, manual information extraction is labor-intensive, resource-inefficient, susceptible to human error, and resistant to scaling methodologies. Given the projected submission of more than 300 hydropower relicensing applications within the next decade, there exists an urgent necessity for an automated solution that offers precision, efficiency, and scalability.

Recent advancements in Natural Language Processing (NLP) demonstrate significant potential for automating complex regulatory document analysis. Transformer-based models (e.g., Bidirectional Encoder Representations from Transformers (BERT) Devlin et al. [2019], Generative Pre-trained Transformer (GPT)) have enhanced capabilities for contextual understanding and structured information extraction from unstructured texts. While these models have proven effective in biomedical and legal domains, their application remains limited in hydropower regulatory compliance contexts.

In our previous research, we developed a comprehensive database of FERC's environmental mitigation requirements, identifying 135 distinct mitigation categories from extensive licensing documentation. This structured dataset facilitated the training of a BERT-based supervised classification model. Building upon this foundation, we present a sophisticated NLP framework designed to extract 17 fundamental information fields from hydropower mitigation documents. Our methodology involved the systematic evaluation of various open-weight LLMs across different parameter sizes, employing multiple prompt-based extraction methodologies for quantitative assessment. Furthermore, we implemented novel reasoning algorithms to optimize extraction accuracy and reliability.

The primary contributions of this research comprise: (1) a comprehensive quantitative analysis of multiple open-weight LLM-based extraction methodologies optimized for domain-specific information retrieval, (2) identification of a critical parameter threshold where LLMs achieve substantial improvements over traditional NLP baselines, and (3) the implementation of novel reasoning strategies that demonstrably enhance extraction performance. These methodological advances contribute to the broader field of NLP applications in regulatory compliance, with significant implications for analogous specialized domains.

## 2 Information extraction: a review

Initial approaches to NLP employed deterministic methodologies based on linguistic rule systems and heuristic algorithmic frameworks. Key implementations including Stanford NLP and Open NLP incorporated computational linguistics techniques such as morphological analysis, named entity recognition (NER), and syntactic parsing for textual data analysis and information extraction. While these rule-based methodologies demonstrated high precision in structured text analysis, their performance metrics exhibited significant degradation when processing non-standard linguistic patterns, syntactic irregularities, orthographic variations, and stochastic noise in input data. This limited robustness to linguistic variation constituted a fundamental constraint in conventional NLP architectures. Moreover, rule-based implementations necessitated extensive manual feature engineering and demonstrated poor computational scalability, limiting their utility for high-throughput text processing applications requiring sophisticated linguistic analysis capabilities.

The advent of Deep Learning (DL) methodologies revolutionized NLP through the implementation of neural architectures capable of encoding complex linguistic patterns and semantic representations. Text Convolutional Neural Networks (Text CNNs), as pioneered by Kim Kim [2014], exhibited superior performance metrics across multiple NLP tasks, including sentiment classification, topic modeling, and document categorization, through the application of convolutional operators for local feature extraction. The subsequent development of Recurrent Neural Networks (RNNs) and their architectural variants, specifically Long Short-Term Memory networks (LSTMs), enhanced sequential text processing capabilities through temporal modeling frameworks. These architectures demonstrated robust performance in encoding sequential dependencies, enabling significant advancements in neural machine translation, probabilistic language modeling, automated speech recognition, and entity extraction systems, as demonstrated in seminal publications by Hochreiter and Schmidhuber Hochreiter and Schmidhuber [1997], Graves Graves et al. [2013], and Bahdanau et al. Bahdanau et al. [2014]. However, RNN and LSTM architectures exhibited computational limitations in processing extended temporal dependencies, primarily due to gradient instability phenomena and computational complexity constraints, necessitating the development of alternative architectural paradigms.

The proliferation and implementation of LLMs, particularly GPT-3 and its successor GPT-4 Brown et al. [2020], OpenAI [2023], have established new performance benchmarks in natural language processing. Contemporary LLMs exhibit superior capabilities in contextual analysis, logical reasoning, and information extraction methodologies. These computational models demonstrate exceptional efficacy in processing linguistic variations and converting unstructured, domain-specific textual data into structured formats, achieving performance metrics that substantially exceed previous methodological approaches. The architectural flexibility and generalization capacity of LLMs have positioned them as predominant frameworks in NLP research and applications, encompassing regulatory documentation analysis, legal corpus interpretation, biomedical data extraction, and environmental compliance assessment. This methodological progression from deterministic rule-based architectures through neural network implementations to Transformer-based LLMs represents the systematic advancement in NLP frameworks. These technological developments provide robust computational solutions for increasingly sophisticated domain-specific NLP implementations.

Recent investigations into large language model deployment have identified critical reliability challenges, particularly the phenomenon of hallucination wherein models generate linguistically coherent yet factually inaccurate information Ji et al. [2023], Rawte et al. [2023]. This phenomenon exhibits increased severity in domain-specific applications where precision requirements are paramount, such as legal document analysis, medical record processing, and regulatory compliance assessment. Zhang et al. Zhang et al. [2023] demonstrated that hallucination rates increase significantly when models encounter specialized terminology or complex structured data, presenting substantial obstacles for automated information extraction in technical documentation. Furthermore, Mundler et al Mündler et al. [2023] revealed that traditional evaluation metrics may inadequately capture hallucination phenomena, particularly in scenarios where models exhibit high recall values while simultaneously propagating systematic errors throughout extracted information.

Empirical research into model scaling behaviors has revealed non-linear performance improvements with increasing parameter counts, suggesting the existence of discrete capability emergence thresholds rather than continuous enhancement patterns Wei et al. [2022a], Srivastava [2022]. The "scaling laws" literature demonstrates that certain complex reasoning capabilities emerge only above specific parameter thresholds, with substantial performance discontinuities observed across various NLP benchmarks Kaplan et al. [2020], Hoffmann et al. [2022]. However, existing scaling studies have predominantly focused on general-purpose language tasks, with limited investigation into parameter requirements for specialized information extraction applications. Recent work by Chowdhery et al. Chowdhery et al. [2022] suggests that domain-specific tasks may exhibit distinct scaling patterns compared to general language understanding, necessitating targeted empirical analysis for regulatory and technical document processing applications.

Contemporary research has demonstrated substantial improvements in LLM reliability through structured reasoning methodologies, particularly chain-of-thought prompting techniques that encourage explicit intermediate reasoning steps Wei et al. [2022b], Kojima et al. [2022]. Self-verification approaches, wherein models critically evaluate their own outputs through iterative refinement processes, have shown significant promise in reducing hallucination rates and improving factual accuracy Weng et al. [2023], Dhuliawala et al. [2023]. Recent implementations of multi-step reasoning frameworks, including self-consistency methods and verification-based approaches, have achieved notable performance improvements in mathematical reasoning, logical inference, and structured data extraction tasks Wang et al. [2023]. However, the application of these reasoning strategies to domain-specific information extraction from regulatory documentation remains largely unexplored, presenting opportunities for methodological advancement in specialized NLP applications.

## 3 Methods

### 3.1 Dataset

To evaluate our information extraction framework's efficacy, we analyzed a dataset of FERC hydropower licensing documents obtained from the publicly accessible FERC eLibrary. The corpus comprised licenses issued between 2014-2017 containing environmental mitigation requirements. A significant methodological challenge encountered was the non-standardized distribution of mitigation information throughout these documents. Rather than being consolidated in designated sections, relevant data was dispersed across multiple document components, substantially increasing extraction complexity.

For each year within this temporal range, two licensing documents were randomly selected. The sampling procedure was conducted without replacement and strategically constrained to ensure comprehensive coverage across diverse geographic and operational contexts. Each document encompasses multiple pages and contains extensive environmental, operational, and technical specifications pertinent to the respective hydropower project.

Data preprocessing for compatibility with LLMs involved segmentation of each document into manageable textual units. Considering the token length constraints inherent to most LLMs and the imperative to maintain contextual integrity for optimal information extraction, documents were systematically divided into chunks of 1,000 words. To mitigate information loss at segmentation boundaries, an overlap of 200 words was implemented between adjacent chunks. This methodological approach preserved semantic continuity and minimized the potential fragmentation of contextually relevant information.

The preprocessing protocol yielded 209 document chunks, which functioned as the primary input units for subsequent entity extraction experiments, facilitating rigorous evaluation of model performance across heterogeneous and ecologically valid document structures.

### 3.2 Model selection

This investigation evaluated seven open-weight large language models with parameter counts ranging from 0.6B to 70B, strategically selected to examine scaling effects and identify critical performance thresholds in information extraction tasks. The selection encompassed three distinct parameter regimes based on established scaling law literature, complemented by a traditional NLP baseline for comprehensive performance comparison.

**Small-scale models (0.6B-3B):** Qwen 0.6B, Llama 1B, and Llama 3B represented computationally efficient architectures suitable for edge deployment scenarios. These models facilitated assessment of baseline extraction capabilities and validation of the hypothesis that sub-threshold models exhibit systematic hallucination patterns.

**Mid-scale model (14B):** Qwen 14B constituted a critical evaluation point based on emerging evidence suggesting capability transitions around this parameter scale. While scaling law research indicates performance improvements correlate with increased parameters Kaplan et al. [2020], Brown et al. [2020], the 14B parameter range presents an optimal experimental context to quantitatively assess whether meta-cognitive validation mechanisms demonstrate enhanced efficacy, thereby establishing empirical benchmarks for threshold identification.

**Large-scale models (20B-70B):** GPT-OSS 20B, Qwen 32B, and Llama 70B represented state-of-the-art parameter scales, enabling comprehensive assessment of scaling benefits and identification of performance plateaus. These models tested whether emergent meta-cognitive capabilities continue improving beyond the initial threshold.

**Traditional baseline:** A SpaCy Honnibal and Montani [2017]-based named entity recognition system combined with domain-specific rule-based patterns provided a non-neural baseline, establishing performance benchmarks that large language models must surpass to justify their computational overhead in regulatory document processing applications.

Selection criteria prioritized: (1) open-weight availability ensuring methodological transparency and reproducibility, (2) instruction-tuned variants optimized for task adherence, (3) diverse architectural paradigms spanning Transformer variants and training methodologies, and (4) strategic coverage of documented parameter thresholds in scaling research Wei et al. [2022a].

All models were accessed through Hugging Face's model hub and deployed using standardized inference parameters to ensure methodological consistency. Proprietary models were deliberately excluded to maintain complete transparency and reproducibility of the experimental methodology, enabling full replication by the research community.

### 3.3 Information fields for extraction

This study aims to evaluate the efficacy of LLMs in extracting structured information from unstructured regulatory documentation. We identified 17 target fields encompassing critical hydropower project parameters and environmental management metrics. These fields were systematically derived through comprehensive analysis of hydropower licensing documentation and represent essential data elements required for regulatory compliance assessment, environmental monitoring protocols, and infrastructure planning frameworks.

To enhance analytical clarity and optimize model performance, we categorized these 17 fields into six functional classifications:

**Basic Information (4 fields)**

- **Dam_Name**: The official name of the dam, structure, or project.
- **Location**: The general location of the project, including city, state, and river.
- **County**: The name(s) of the county or counties in which the project is located.
- **Primary_Purpose**: The principal function of the project, such as hydropower generation, flood control, irrigation, or recreation.

**Flow Information (4 fields)**

- **Minimum_Flow**: The mandated minimum flow rate (typically in cubic feet per second, cfs) that must be maintained.
- **Annual_Flow_Peak**: The maximum flow recorded over the course of a year (cfs).
- **Annual_Flow_Mean**: The mean annual flow rate (cfs).
- **Spillway_Maximum_Discharge_Flow**: The maximum discharge capacity of the spillway (cfs).

**Elevation Information (5 fields)**

- **Maximum_Pool_Elevation**: The highest allowable elevation of the pool or reservoir (in feet).
- **Normal_Maximum_Operating_Pool_Level**: The normal upper operating level under standard conditions (feet).
- **Maximum_Operating_Pool_Level**: The highest elevation at which the project may operate (feet).
- **Minimum_Pool_Elevation**: The lowest allowable pool elevation (feet).
- **Power_Head**: The effective vertical drop (head) driving the turbines (feet).

**Capacity Information (2 fields)**

- **Power_Capacity**: The total installed generating capacity of the facility (in megawatts, MW).
- **Energy_Output**: The annual energy output or generation (in megawatt-hours, MWh).

**Storage Information (1 field)**

- **Usable_Storage_Volume**: The volume of reservoir storage that is actively used for power generation or other operational purposes (acre-feet).

**Environmental Information (1 field)**

- **Stream_Temperature**: Information related to stream temperature targets, thresholds, or control strategies.

These field definitions constituted the reference standard for quantitative assessment of model extraction performance metrics and informed the systematic development of prompt engineering methodologies within our LLM-based analytical framework.

### 3.4 Information extraction methods

We evaluated four distinct prompting strategies for extracting hydropower information from regulatory documents. Each strategy represents a methodological approach to information extraction, designed to explore various prompting paradigms and their effectiveness across different model architectures.

Scaling Open-Weight LLMs for Hydropower Regulatory Information Extraction     A PREPRINT### 3.4.1 Single-step extraction

The single-step method employs a direct extraction approach where all 17 fields are requested simultaneously in a single prompt. This baseline method presents the most straightforward task structure, instructing the model to identify and extract all required information in one computational pass. The prompt explicitly lists each field with its description and requests the output in a standardized JSON format, with null values for absent information.

This method minimizes computational overhead by presenting a simple field-to-value mapping task without intermediate reasoning steps. The approach relies on the model's ability to comprehend the document holistically and directly map textual information to the appropriate fields. While computationally efficient, this method may be susceptible to hallucination when models attempt to populate fields without sufficient textual evidence.

### 3.4.2 Two-step extraction

The two-step method implements a verification-then-extraction pipeline designed to reduce false positive extractions. In the first phase, the model evaluates the presence of information for each field, responding with "YES" if clearly present, "NO" if absent, or "MAYBE" if uncertain. Only fields marked as "YES" or "MAYBE" proceed to the second phase for actual value extraction.

This approach introduces a structured decision-making process that requires the model to explicitly consider information availability before attempting extraction. By separating existence verification from value extraction, the method aims to reduce hallucination and improve extraction precision. However, this multi-turn approach requires models to maintain context across sequential prompts and may introduce failure modes in models with limited context retention capabilities.

### 3.4.3 Categorical extraction

The categorical method organizes the 17 extraction fields into six semantic groups: Basic Information (dam name, location, county, primary purpose), Flow Information (minimum flow, annual flow statistics, spillway discharge), Elevation Information (pool elevations, power head), Capacity Information (power capacity, energy output), Storage Information (usable storage volume), and Environmental Information (stream temperature). This hierarchical organization provides cognitive scaffolding that mirrors the natural structure of hydropower documentation.

By grouping related fields, this method enhances contextual understanding and facilitates models in identifying relevant information more effectively. The semantic clustering creates natural search boundaries within the document, potentially reducing cognitive load while maintaining comprehensive coverage.

### 3.4.4 Chain-of-Thought extraction

The chain-of-thought method guides models through an explicit six-step reasoning process before extraction. The steps include: (1) document overview and type identification, (2) basic project information search, (3) flow and water management analysis, (4) physical infrastructure identification, (5) power generation details extraction, and (6) environmental considerations review. Only after completing this systematic analysis does the model proceed to structured field extraction.

This approach leverages the chain-of-thought prompting paradigm to encourage thorough document analysis and reduce oversight errors. By explicitly outlining the reasoning pathway, the method aims to improve extraction accuracy through systematic attention to different document sections.

### 3.4.5 Implementation details

All extraction methods were implemented with consistent parameters to ensure fair comparison. We used deterministic generation (temperature = 0) to enhance reproducibility, allocated sufficient tokens for complete JSON responses, and maintained identical input formatting across all methods. Each method processed the same document chunks and produced outputs in a standardized JSON structure containing all 17 fields, with null values indicating absent or unextractable information.

The diversity of these extraction approaches allows for comprehensive evaluation of model capabilities across different prompting paradigms, providing insights into optimal strategies for various model scales and use cases. Our experimental design enables direct comparison of cognitive load, extraction accuracy, and computational efficiency across methods.

## 3.5 Reflective reasoning for quality assurance

Beyond the four primary extraction methods, we introduce a novel reasoning paradigm called reflective reasoning, which applies post-extraction validation to assess and improve the quality of extracted information. Unlike forward reasoning approaches that guide the extraction process, reflective reasoning retrospectively evaluates extraction outputs against source text, implementing a metacognitive verification process that mirrors human quality assurance practices.

### 3.5.1 Conceptual framework

Reflective reasoning operates on the principle that language models can effectively evaluate their own outputs when provided with appropriate criteria and the original context. This approach leverages the model's ability to perform comparative analysis between extracted values and source text, identifying potential hallucinations, inaccuracies, or unsupported inferences. By separating the extraction and validation phases, reflective reasoning allows for targeted quality control without constraining the initial extraction process.

The framework addresses a critical challenge in LLM-based information extraction: the tendency to generate plausible but unsupported information. Through systematic post-hoc evaluation, reflective reasoning provides a mechanism to filter extraction outputs based on configurable strictness criteria, enabling researchers to optimize for different precision-recall trade-offs. In this study, we applied reflective reasoning validation protocols to the outputs generated through the single-step extraction methodology to quantitatively assess its efficacy as a quality enhancement mechanism.

### 3.5.2 Implementation strategies

We implemented three distinct validation strategies within the reflective reasoning framework, each representing a different point on the precision-recall spectrum:

**Lenient validation**    The lenient validation strategy implements permissive reflective reasoning with a bias toward accepting extracted values. This approach prioritizes high recall by minimizing false negatives, accepting values that appear in the text or can be reasonably inferred from context. When evidence exists but is not explicit, the validation leans toward acceptance. The method targets moderate rejection rates to preserve comprehensive information capture while filtering only clearly erroneous extractions. This strategy targets low rejection rates of 20-30% to preserve comprehensive information capture while filtering only clearly erroneous extractions.

**Stringent validation**    The stringent validation strategy implements restrictive reflective reasoning requiring exact textual support for all extracted values. This approach demands word-for-word correspondence between extracted values and source text, rejecting any extraction requiring inference, paraphrasing, or format transformation. For numerical fields, even minor variations such as rounding or unit abbreviation trigger rejection. The method incorporates sophisticated uncertainty detection, automatically rejecting responses containing qualifying language. This strategy aims to maximize precision through aggressive filtering. This strategy aims for high rejection rates of 60-80% to maximize precision through aggressive filtering.

**Moderate validation**    The moderate validation strategy implements adaptive reflective reasoning that seeks optimal overall performance. This approach accepts reasonable variations while maintaining accuracy standards, permitting standard abbreviations and format differences while requiring exact numerical values. The validation applies practical judgment criteria, accepting essential information matches while rejecting clear errors. This strategy targets a middle ground between the lenient and stringent approaches. This strategy targets moderate rejection rates of 40-60% to achieve balanced precision-recall performance.

## 3.6 Bronze standard

While gold standard annotations created by human experts represent the ideal evaluation benchmark for information extraction tasks, the practical constraints of expert annotation—including time, cost, and the specialized domain knowledge required for technical documents—necessitated an alternative approach. We adopted an LLM-as-a-Judge methodology to create what we term a "Bronze Standard" evaluation dataset.

Bronze standard annotations may contain inherent errors and biases associated with the judging model. However, for comparative analysis of extraction methods and model performance, this approach provides consistent and scalable evaluation criteria. The bronze standard facilitates systematic comparison of relative performance across different configurations, revealing trends and patterns that remain valid despite potential absolute accuracy limitations.

We generated the bronze standard using GPT-4o mini as an evaluator, leveraging its demonstrated performance on judgment tasks and semantic understanding capabilities. For each field in each document chunk, we presented the evaluator with the source text and instructed it to identify and extract the correct value if present. The evaluation prompt emphasized conservative extraction—only identifying values with explicit textual support and returning null values for absent or ambiguous information.

To ensure methodological rigor, we implemented an iterative refinement process. Initial bronze standard creation revealed excessive rejection rates that would prevent statistically meaningful model comparison. Through systematic adjustment of the evaluation parameters, we developed a balanced methodological approach that maintains data integrity while preserving sufficient positive examples for robust statistical evaluation.

### 3.7 Evaluation methodology

The bronze standard evaluation implements semantic comparison rather than exact string matching, recognizing that valid extractions may exhibit format variations while preserving semantic equivalence. For example, "100 MW" and "100 megawatts" represent functionally equivalent extractions despite surface-level textual differences. This semantic evaluation methodology more accurately reflects real-world information extraction requirements where semantic accuracy supersedes formatting precision.

Our evaluation framework compares model outputs against bronze standard annotations using precision, recall, and F1 metrics calculated at the field level across all document chunks. This granular evaluation methodology enables detailed statistical analysis of model performance across diverse field types and extraction challenges.

### 3.8 Experimental setup

All experiments utilized deterministic generation parameters to ensure reproducibility. Temperature was fixed at 0 across all model inferences, thereby eliminating sampling variability and facilitating consistent outputs across multiple experimental runs. Maximum token thresholds were calibrated to accommodate complete JSON response structures for all 17 target fields, with appropriate buffer allocation to prevent response truncation.

Models were executed on GPU infrastructure with optimized memory allocation protocols for large language model inference. Batch processing methodologies were implemented with precisely calibrated batch dimensions to maximize computational throughput while circumventing memory limitations, employing reduced batch parameters for higher-parameter models and validation procedures.

Each model-method permutation was executed once on the comprehensive dataset comprising 209 document segments, generating 49 distinct experimental configurations (7 models × 7 methodologies). For reflective reasoning validation protocols, we applied three distinct validation strategies exclusively to single-step extraction outputs, establishing an additional evaluation dimension. All experimental results were preserved in structured JSON format to facilitate comprehensive post-hoc analytical procedures and ensure complete reproducibility of evaluation metrics.

## 4 Results

We evaluated seven language models across seven different information extraction approaches, yielding 49 distinct experimental configurations. Performance assessment utilized precision, recall, and F1 metrics computed at the field level across 209 document chunks containing hydropower regulatory information.

### 4.1 Overall performance

Table 1 presents comprehensive F1 scores for all model-method combinations. Llama 70B with stringent verification exhibited maximum performance (F1 = 0.767), followed by Llama 70B with moderate verification (F1 = 0.736). Among large-scale models, Qwen 32B demonstrated consistent performance across all verification methods (F1 = 0.742-0.766), while GPT-OSS 20B yielded optimal results with categorical extraction (F1 = 0.669).

Ranked by F1 score, the highest-performing model-method configurations were: (1) Llama 70B with Stringent verification (0.767), (2) Qwen 32B with Lenient verification (0.766), (3) Qwen 32B with Moderate verification (0.764), and (4) Qwen 32B with Stringent verification (0.742).

Precision values exhibited significant variation across models and methods, ranging from 0.015 to 0.853. GPT-OSS 20B demonstrated exceptional precision in moderate verification (0.853) and stringent verification (0.731), indicating conservative but accurate extraction behavior. Large-scale models exhibited balanced precision performance, with

Table 1: F1 Score Results

| Model | Extraction Methods | | | | Reflective Reasoning | | |
|---|---|---|---|---|---|---|---|
| | Single-step | Two-step | Categorical | Chain-of-Thought | Lenient | Moderate | Stringent |
| Qwen 0.6B | 0.348 | 0.262 | 0.420 | 0.397 | 0.066 | 0.079 | 0.066 |
| Llama 1B | 0.318 | 0.277 | 0.318 | 0.319 | 0.090 | 0.067 | 0.077 |
| Llama 3B | 0.344 | 0.000 | 0.508 | 0.363 | 0.118 | 0.132 | 0.102 |
| Qwen 14B | 0.551 | 0.235 | 0.495 | 0.532 | 0.513 | 0.633 | 0.640 |
| GPT-OSS 20B | 0.630 | 0.041 | 0.669 | 0.553 | 0.631 | 0.515 | 0.608 |
| Qwen 32B | 0.767 | 0.048 | 0.636 | 0.344 | 0.766 | 0.764 | 0.742 |
| Llama 70B | 0.702 | 0.051 | 0.711 | 0.691 | 0.701 | 0.736 | 0.767 |
| Traditional NLP | 0.549 | – | – | – | – | – | – |

Table 2: Precision Results

| Model | Extraction Methods | | | | Reflective Reasoning | | |
|---|---|---|---|---|---|---|---|
| | Single-step | Two-step | Categorical | Chain-of-Thought | Lenient | Moderate | Stringent |
| Qwen 0.6B | 0.211 | 0.151 | 0.272 | 0.253 | 0.034 | 0.041 | 0.034 |
| Llama 1B | 0.189 | 0.160 | 0.190 | 0.190 | 0.047 | 0.035 | 0.040 |
| Llama 3B | 0.210 | 0.000 | 0.432 | 0.242 | 0.063 | 0.071 | 0.054 |
| Qwen 14B | 0.629 | 0.134 | 0.593 | 0.578 | 0.345 | 0.463 | 0.471 |
| GPT-OSS 20B | 0.697 | 0.024 | 0.685 | 0.644 | 0.704 | 0.853 | 0.731 |
| Qwen 32B | 0.691 | 0.025 | 0.621 | 0.506 | 0.702 | 0.698 | 0.689 |
| Llama 70B | 0.586 | 0.029 | 0.592 | 0.569 | 0.586 | 0.775 | 0.691 |
| Traditional NLP | 0.419 | – | – | – | – | – | – |

Llama 70B achieving 0.775 precision in moderate verification and Qwen 32B maintaining consistent precision ranging from 0.689-0.702 across all verification approaches.

Table 3: Recall Results

| Model | Extraction Methods | | | | Reflective Reasoning | | |
|---|---|---|---|---|---|---|---|
| | Single-step | Two-step | Categorical | Chain-of-Thought | Lenient | Moderate | Stringent |
| Qwen 0.6B | 0.975 | 1.000 | 0.929 | 0.926 | 1.000 | 1.000 | 1.000 |
| Llama 1B | 0.987 | 1.000 | 0.976 | 1.000 | 1.000 | 1.000 | 1.000 |
| Llama 3B | 0.950 | 0.000 | 0.616 | 0.723 | 1.000 | 1.000 | 1.000 |
| Qwen 14B | 0.490 | 0.994 | 0.424 | 0.493 | 1.000 | 1.000 | 1.000 |
| GPT-OSS 20B | 0.575 | 0.149 | 0.654 | 0.485 | 0.572 | 0.368 | 0.520 |
| Qwen 32B | 0.862 | 0.994 | 0.652 | 0.268 | 0.848 | 0.850 | 0.798 |
| Llama 70B | 0.875 | 0.224 | 0.888 | 0.879 | 0.873 | 0.702 | 0.862 |
| Traditional NLP | 0.795 | – | – | – | – | – | – |

Recall analysis revealed distinct behavioral patterns across model scales. Small-scale models (0.6B-3B) consistently achieved perfect recall scores (1.000) during verification procedures, indicative of systematic over-extraction phenomena characteristic of hallucination mechanisms. In contrast, large-scale architectures demonstrated statistically differentiated recall distributions: Llama 70B (0.702-0.873), Qwen 32B (0.796), and GPT-OSS 20B (0.368-0.572). These graduated recall metrics suggest enhanced discriminative capacity between extractable and non-extractable information domains, reflecting more sophisticated information boundary recognition.

## 4.2   Performance by model size

A clear performance threshold emerges at 14B parameters. Models below 14B achieved maximum F1 scores of 0.508 (Llama 3B categorical extraction), while all models at 14B and above exceeded 0.513 F1 performance in their best configurations, representing a 26% absolute improvement at the threshold. For extraction methods, models ≥14B demonstrated F1 scores ranging from 0.235 to 0.767, whereas small models (0.6B-3B) exhibited scores of 0.000 to 0.508. This performance differential was significantly amplified in reflective reasoning methodologies, where large models attained F1 scores between 0.513 and 0.767, substantially exceeding the 0.066 to 0.132 range observed in small models.

### 4.3 Performance by method

Categorical extraction demonstrated superior efficacy, yielding optimal performance in two of four models (Qwen 0.6B: 0.420, Llama 3B: 0.508). Single-step extraction exhibited consistent baseline performance across all model architectures. Two-step extraction manifested systematically poor performance across all model scales, with complete methodological failure in Llama 3B (F1 = 0.000) and severe performance degradation in large-scale models: Qwen 32B (F1 = 0.048), Llama 70B (F1 = 0.051), and GPT-OSS 20B (F1 = 0.041). Within reflective reasoning approaches, stringent validation protocols produced the highest individual F1 metric (0.767 for Llama 70B) but exhibited heterogeneous performance across different model implementations.

### 4.4 Performance by information type

To investigate the relationship between information complexity and extraction performance, we analyzed F1 scores across six distinct information categories using single-step extraction results. Table 4 presents category-specific performance metrics using macro-averaged F1 scores derived from field-level measurements with semantic matching evaluation.

Table 4: Category-wise Performance Analysis (Single-step Extraction)

(a) F1 Score

| Model | Basic | Flow | Elevation | Capacity | Storage | Environment |
|---|---|---|---|---|---|---|
| Qwen 0.6B | 0.205 | 0.045 | 0.065 | 0.087 | 0.052 | 0.000 |
| Llama 1B | 0.282 | 0.029 | 0.063 | 0.101 | 0.041 | 0.000 |
| Llama 3B | 0.235 | 0.068 | 0.055 | 0.082 | 0.036 | 0.010 |
| Qwen 14B | 0.526 | 0.696 | 0.602 | 0.774 | 0.625 | 0.000 |
| GPT-OSS 20B | 0.486 | 0.667 | 0.481 | 0.753 | 0.600 | 0.190 |
| Qwen 32B | 0.517 | 0.634 | 0.469 | 0.812 | 0.833 | 0.091 |
| Llama 70B | 0.486 | 0.723 | 0.647 | 0.804 | 0.769 | 0.308 |
| Traditional NLP | 0.114 | 0.575 | 0.167 | 0.659 | 1.000 | 0.000 |

(b) Precision

| Model | Basic | Flow | Elevation | Capacity | Storage | Environment |
|---|---|---|---|---|---|---|
| Qwen 0.6B | 0.114 | 0.023 | 0.034 | 0.046 | 0.027 | 0.000 |
| Llama 1B | 0.164 | 0.015 | 0.033 | 0.054 | 0.021 | 0.000 |
| Llama 3B | 0.133 | 0.035 | 0.028 | 0.043 | 0.019 | 0.005 |
| Qwen 14B | 0.375 | 0.706 | 0.519 | 0.706 | 0.455 | 0.000 |
| GPT-OSS 20B | 0.461 | 0.909 | 0.475 | 0.762 | 0.600 | 0.182 |
| Qwen 32B | 0.372 | 0.578 | 0.390 | 0.736 | 0.714 | 0.062 |
| Llama 70B | 0.329 | 0.667 | 0.524 | 0.709 | 0.625 | 0.267 |
| Traditional NLP | 0.065 | 0.460 | 0.158 | 0.545 | 1.000 | 0.000 |

(c) Recall

| Model | Basic | Flow | Elevation | Capacity | Storage | Environment |
|---|---|---|---|---|---|---|
| Qwen 0.6B | 0.978 | 0.850 | 0.969 | 0.895 | 1.000 | 0.000 |
| Llama 1B | 1.000 | 0.923 | 1.000 | 0.957 | 1.000 | 0.000 |
| Llama 3B | 0.965 | 0.958 | 1.000 | 0.889 | 1.000 | 1.000 |
| Qwen 14B | 0.878 | 0.686 | 0.718 | 0.857 | 1.000 | 0.000 |
| GPT-OSS 20B | 0.513 | 0.526 | 0.487 | 0.744 | 0.600 | 0.200 |
| Qwen 32B | 0.843 | 0.703 | 0.590 | 0.907 | 1.000 | 0.167 |
| Llama 70B | 0.926 | 0.789 | 0.846 | 0.929 | 1.000 | 0.364 |
| Traditional NLP | 0.441 | 0.767 | 0.176 | 0.833 | 1.000 | 0.000 |

Capacity Information exhibited the highest performance across all models (average F1: 0.509), comprising technical specifications including power capacity and energy output that benefit from standardized units and clear numerical formatting. Storage Information demonstrated comparable performance (average F1: 0.495) despite being a single field, indicating the effectiveness of pattern recognition for well-defined storage volume metrics.

Flow Information achieved moderate performance (average F1: 0.429), with substantial variation across models. Large-scale models (14B+) demonstrated competent extraction (F1: 0.634-0.723), while smaller models exhibited systematic failures (F1: 0.029-0.068), particularly for Annual_Flow_Mean and Annual_Flow_Peak fields.

Environmental Information presented the greatest extraction challenge across all model architectures (average F1: 0.075), with complete extraction failure in most small-scale models and only modest success in large-scale implementations. This systematic limitation reflects the highly specialized and context-dependent nature of stream temperature regulations within hydropower documentation.

The observed performance hierarchy reveals that technical specifications with standardized formatting outperform contextual identifiers, contrasting with conventional assumptions about textual versus numerical extraction difficulty. This pattern suggests that semantic complexity and regulatory specificity constitute more significant factors than data type in determining extraction success.

### 4.5 Validation strategy performance

Table 5 presents the rejection rates observed for each validation strategy relative to the predetermined target range of 40-60

Table 5: Validation Rejection Rates

| Model      | Lenient | Moderate | Stringent | Target Achievement    |
|------------|---------|----------|-----------|-----------------------|
| Qwen 0.6B  | 1.6%    | 27.0%    | 40.6%     | Stringent only        |
| Llama 1B   | 5.6%    | 61.6%    | 84.1%     | None                  |
| Llama 3B   | 5.5%    | 46.8%    | 89.0%     | Moderate only         |
| Qwen 14B   | 16.8%   | 53.4%    | 91.2%     | Moderate only         |
| GPT-OSS 20B| 29.6%   | 48.5%    | 63.2%     | Moderate only         |
| Qwen 32B   | 30.8%   | 30.8%    | 30.8%     | None (convergence)    |
| Llama 70B  | 0.1%    | 3.2%     | 7.5%      | None (low rejection)  |

Validation rejection rates exhibited significant variability across models and validation strategies (Table 5). Lenient validation consistently yielded minimal rejection rates (1.6%-16.8%) across all tested models. In contrast, stringent validation produced substantially higher rejection rates in 85.7% of models, with rejection frequencies ranging from 40.6% to 91.2%. Moderate validation generated intermediate rejection rates (27.0%-61.6%), with 57% of models achieving values within the predetermined 40-60% optimal threshold.

Large-parameter models demonstrated distinctive validation characteristics. Qwen 32B maintained uniform rejection rates (30.8%) across all validation methodologies, suggesting convergent decision-making processes regardless of stringency parameters. Llama 70B exhibited consistently low rejection frequencies (0.1%-7.5%) across all validation conditions, indicating minimal discriminative capability. GPT-OSS 20B displayed clear stratification between validation approaches, with rejection rates of 29.6%, 48.5%, and 63.2% for lenient, moderate, and stringent validation protocols, respectively.

The target rejection range (40-60%) was achieved by specific model-validation pairings: Qwen 0.6B (stringent), Llama 3B (moderate), Qwen 14B (moderate), and GPT-OSS 20B (moderate). Notably, 42.9% of models failed to achieve optimal rejection rates under any validation condition.

### 4.6 Notable findings

Several noteworthy anomalies emerged from the analysis. The Llama 3B model with two-step extraction yielded F1, precision, and recall scores of 0.000, demonstrating complete extraction failure. This configuration represents the sole instance where no valid extractions were produced across the entire corpus of 209 document chunks.

Small-scale models (0.6B-14B) maintained perfect recall (1.0000) in reflective reasoning methodologies, while large-scale models (20B-70B) demonstrated more realistic recall values (0.368-0.873), suggesting enhanced discriminative capacity. Precision exhibited significant variability across all models (0.0340-0.8530). This performance pattern suggests that validation approaches successfully preserved all true positive extractions while demonstrating differential efficacy in false positive filtration.

Optimal methodological selection demonstrated model-dependence: categorical extraction yielded superior performance for smaller-scale models (0.6B, 3B), whereas stringent validation protocols achieved maximal efficacy for Qwen 14B. Single-step extraction methodology provided the most consistent baseline performance across the parametric spectrum.

## 5 Discussion

Our investigation of language model-based information extraction from hydropower regulatory documentation identifies a significant paradox: despite enhanced F1 metrics in larger models, all evaluated systems demonstrate abnormally elevated recall values—a phenomenon indicative of hallucination prevalence rather than extraction proficiency. The variable efficacy of reflective reasoning mechanisms—yielding performance enhancements exclusively in the 14B parameter architecture while demonstrating performance degradation in smaller-scale models—suggests a quantifiable threshold for meta-cognitive functionality within neural language systems. These empirical observations have substantial implications for language model implementation within critical regulatory compliance frameworks where extraction errors may precipitate consequential outcomes.

### 5.1 The hallucination hypothesis: a signal detection framework

We propose conceptualizing LLM hallucination through a signal detection theory framework that distinguishes between "signal exists" and "signal does not exist" conditions. Traditional machine learning models demonstrate proficiency in identifying positive signals—detecting patterns when distinctive features are present. However, they systematically underperform when recognizing negative signals—confidently asserting information absence—because the absence of information lacks distinguishing features, manifesting merely as baseline noise distributions.

This asymmetry manifests in LLMs as a fundamental architectural bias: these models are optimized to generate probabilistically likely continuations given input sequences, with limited mechanisms for uncertainty quantification Xiong et al. [2024]. When confronted with absent information, LLMs default to generating plausible values rather than acknowledging epistemic uncertainty, resulting in the elevated recall rates observed across all models (0.490-1.000 for extraction methods). Recent empirical evaluations by Vectara's Hallucination Evaluation Model found substantial variation in hallucination rates across different LLMs, with rates ranging from 3% for OpenAI's models to 27% for Google's PaLM-Chat Metz [2023]. These findings underscore the persistent challenge of hallucination in current LLM architectures, particularly when models encounter domains with absent or ambiguous information.

Three convergent patterns in our experimental results support the hallucination hypothesis. First, universally high recall metrics: small parameter models (0.6B-3B) achieved recall rates of 0.929-1.000 despite demonstrating poor precision (0.151-0.432), suggesting systematic extraction of values even when information is objectively absent. This pattern aligns with recent findings from ACL 2024 demonstrating that LLMs achieve "satisfactory recall on unstructured attributes with suboptimal precision, primarily due to the inherent ambiguity of unstructured attributes" Jiang et al. [2024].

Second, complete field extraction failures: Annual_Flow_Mean and Annual_Flow_Peak achieved F1 = 0.000 across all model architectures, indicating systematic hallucination when specific information types are consistently absent from source documents. These failures represent archetypal "signal does not exist" cases where models cannot discriminate between genuine information absence and extractable content.

Third, perfect validation recall in small models: small-scale models (0.6B-14B) in reflective reasoning approaches maintained perfect recall (1.000) while precision varied substantially (0.034-0.853), while large-scale models (20B-70B) demonstrated more realistic recall distributions (0.368-0.873). This demonstrates that validation processes in smaller models suffer from identical positive generation bias, revealing what we term "meta-hallucination"—the validation model's inability to recognize extraction hallucinations within its own cognitive framework. Large models show enhanced discriminative capacity but still exhibit systematic biases in certain domains.

### 5.2 Field distribution bias and selective hallucination patterns

Our category-wise analysis reveals a critical limitation in standard evaluation metrics that has significant implications for real-world deployment. While small models (0.6B-3B) achieve seemingly impressive overall recall scores (0.929-1.000), this performance masks severe category-specific failures that pose substantial regulatory compliance risks.

The observed pattern challenges conventional assumptions about uniform hallucination in small language models. Rather than blanket over-extraction, these models demonstrate selective hallucination: high performance in Basic Information fields (recall: 0.965-1.000) and Storage Information (recall: 1.000 across all small models) while failing systematically in technical categories such as Flow Information (recall: 0.850-0.958) and Environmental compliance (recall: 0.000-1.000 with extreme variability).

This selective pattern occurs because Basic Information, Capacity Information, and Storage Information comprise approximately 41% of all extractable fields (7 out of 17), yet demonstrate the highest extraction success rates. High-frequency, straightforward extractions dominate aggregate metrics while critical but challenging regulatory categories remain poorly served. A model achieving perfect recall (1.000) for storage volume may simultaneously achieve zero recall for environmental compliance—a potentially catastrophic failure mode masked by conventional evaluation approaches.

The 14B parameter threshold exhibits pronounced category-dependent effects across all six categories. Basic Information shows modest improvement from small to large models (0.327 to 0.499 average F1, +52% improvement). However, technical categories demonstrate exponential scaling benefits: Flow Information improves from 0.047 to 0.674 (+1,334% improvement), Capacity Information from 0.090 to 0.780 (+767% improvement), and Elevation Information from 0.061 to 0.532 (+772% improvement). Storage Information shows moderate scaling (0.043 to 0.734, +1,607% improvement), while Environmental Information remains largely intractable across all parameter scales (0.003 to 0.196, +6,433% improvement but still extremely low absolute performance).

This suggests that parameter scaling benefits are highly non-uniform, with reasoning-intensive and context-dependent domains requiring substantially higher computational capacity for effective extraction compared to well-structured textual identifiers.

### 5.3 Two types of reasoning, different mechanisms

Our implementation of two reasoning paradigms—chain-of-thought (forward reasoning) and reflective reasoning (backward validation)—revealed fundamentally different failure modes despite both attempting to mitigate hallucination. Recent research on improving LLM reliability demonstrates that incorporating external knowledge sources during reasoning and enabling models to verify or revise their own outputs can generate more accurate and coherent responses Kumar et al. [2024]. However, our results suggest this effectiveness is highly contingent on model architecture and parameter count, with reasoning-based validation methods showing meaningful benefits only above the 14B parameter threshold.

Chain-of-Thought methodology directs models through systematized extraction procedures, enhancing structural organization and informational coverage. For lower-parameter models, this approach yielded moderate performance improvements (Qwen 0.6B: 0.348→0.397). However, CoT operates concurrently with generation processes, potentially exacerbating hallucination phenomena by incentivizing models to generate synthetic information when confronted with data absence.

Reflective Reasoning implements post-extraction validation protocols, theoretically enabling hallucination filtration subsequent to initial generation. Our empirical findings, however, demonstrate that this approach exhibits meta-hallucination characteristics: the validation system manifests identical positive detection bias, accepting verisimilar but factually inaccurate extractions. This mechanism elucidates the perfect recall phenomenon—validation systems infrequently reject extracted values due to their limited capacity to definitively identify informational absence.

The observed differential efficacy (demonstrating benefit exclusively in Qwen 14B implementations) suggests these reasoning mechanisms necessitate substantial computational capacity to achieve functional effectiveness, consistent with emergent ability thresholds documented in contemporary literature Wei et al. [2022a].

The observed performance disparity between Qwen 14B and smaller models (0.6B-3B) in reflective reasoning effectiveness corresponds with established research on emergent capabilities in large language models. The literature documents that "reasoning abilities are known to be present in 100B+ parameter models" while being "non-existent in models up to 13B parameters" Zhang et al. [2024]. Our 14B model occupies precisely this critical transitional parameter space, which explains its distinctive capacity to derive benefit from validation methodologies.

This threshold phenomenon is empirically evident in our experimental results: smaller-scale models (0.6B-3B) exhibited validation-induced F1 decreases of 70-80%, indicating insufficient capacity for accurate self-evaluation, whereas Qwen 14B demonstrated a 16.2% F1 improvement, suggesting adequate computational resources for meta-cognitive assessment. Comparable parameter-dependent threshold effects have been documented in recent reasoning model advancements, with OpenAI's o1 achieving 83% accuracy on International Mathematics Olympiad qualifying problems compared to GPT-4o's 13% performance OpenAI [2024], demonstrating the emergence of advanced reasoning capabilities in larger-scale architectures.

The rejection rate patterns (Table 5) provide additional evidence for these threshold-dependent reasoning capabilities. Contrary to initial hypotheses, models with higher baseline precision demonstrated elevated rejection rates. Qwen 14B, despite its superior extraction accuracy, maintained substantial rejection rates (53.4% moderate, 91.2% stringent), indicating that effective validation mechanisms require sufficient computational capacity to identify low-confidence extractions that would otherwise compromise performance metrics.

Large-scale models exhibited minimal rejection behavior (Qwen 32B: 30.8% uniform; Llama 70B: 0.1-7.5%), suggesting their extractions possessed sufficient inherent confidence to withstand validation scrutiny. This pattern confirms the theoretical framework proposed by our collaborators: rejection rates must be interpreted relative to baseline extraction quality rather than assessed against predetermined threshold ranges. High-performing architectures necessitate lower rejection frequencies due to the reduced prevalence of extraction errors requiring remediation.

The validation convergence observed in Qwen 32B—identical 30.8% rejection across all three methodological approaches—represents a particularly significant phenomenon, indicating that at sufficient parameter scales, validation methodologies experience diminished discriminative capacity and converge toward a baseline uncertainty threshold intrinsic to the model's architectural constraints.

The observed parameter-scale dependency indicates that effective hallucination mitigation through reasoning mechanisms requires not merely raw computational capacity but emergent meta-cognitive capabilities that materialize only after exceeding specific parameter thresholds.

### 5.4 Information type analysis: systematic failure patterns

Category-wise analysis revealed that information type, rather than simple text-versus-numerical distinction, determines extraction difficulty and hallucination patterns. Capacity Information (average F1: 0.509) demonstrated highest accessibility across all models, attributed to standardized technical specifications with clear numerical units and consistent formatting in regulatory documents. Storage Information exhibited comparable performance (average F1: 0.495) despite comprising only a single field, indicating highly effective pattern recognition for well-defined volume metrics.

Flow Information (average F1: 0.429) presented moderate extraction challenges, with substantial performance variation across model scales. Large-scale models (14B+) achieved competent extraction (F1: 0.634-0.723), while smaller models exhibited systematic failures (F1: 0.029-0.068). Basic Information (average F1: 0.356) and Elevation Information (average F1: 0.319) demonstrated intermediate difficulty levels, with performance heavily dependent on parameter scale rather than inherent field complexity.

Environmental Information constituted the most challenging category (average F1: 0.075), with near-complete extraction failure across all model architectures. This systematic limitation reflects consistent information absence from source documents rather than extraction difficulty, representing prototypical "signal does not exist" cases where models cannot discriminate between genuine information absence and extractable content.

The 14B parameter threshold exhibits pronounced category-dependent scaling effects. Capacity Information improves from 0.090 (small model average) to 0.780 (large model average), representing 8.7x improvement. Flow Information demonstrates even more dramatic scaling: from 0.047 (small models) to 0.674 (large models), indicating 14.3x improvement. Conversely, Environmental Information remains largely intractable across all parameter scales (0.003 to 0.196), suggesting intrinsic architectural limitations rather than insufficient computational capacity.

These systematic category-specific failures confirm our signal detection theoretical framework: models exhibit competence when distinctive features enable positive signal identification but demonstrate persistent hallucination when confronted with consistent information absence. The category-dependent scaling patterns indicate that reasoning-intensive technical domains require substantially higher computational thresholds for effective extraction compared to well-structured identifiers.

### 5.5 Implications for regulatory information extraction

Our findings present significant implications for the implementation of LLMs in regulatory compliance frameworks, with particular emphasis on the deployment risks associated with aggregate performance metrics.

**Critical deployment warning:** Small models may achieve seemingly acceptable overall performance metrics (F1: 0.318-0.508) while failing catastrophically in mission-critical regulatory categories, posing severe compliance risks. Environmental Information extraction, essential for regulatory compliance, demonstrates near-zero performance (F1: 0.000-0.010) across all small models despite these systems achieving high overall recall (0.929-1.000). A hydropower facility operator relying on small model extraction could systematically miss critical environmental requirements—such as stream temperature monitoring protocols or fish passage specifications—while receiving falsely reassuring aggregate performance indicators. This scenario presents substantial regulatory violation risks that could result in facility shutdown, environmental penalties, or legal liability.

The field distribution bias exacerbates this risk: Basic Information and Capacity fields comprise 41% of extraction targets and demonstrate high success rates, allowing these straightforward extractions to mask systematic failures in specialized regulatory domains. Organizations may incorrectly conclude that a model achieving 75% overall precision is suitable for regulatory deployment, unaware that environmental compliance extraction operates at effectively zero reliability.

**Scale-dependent reliability thresholds:** Reasoning augmentation strategies demonstrate pronounced model-dependent efficacy, requiring parameter thresholds (approximately 14B+) that may exceed computational feasibility in numerous deployment contexts. Organizations must conduct rigorous cost-benefit analyses regarding the relationship between model capacity and reasoning functionality, recognizing that certain regulatory categories remain intractable below critical parameter scales.

**Hybrid system architecture necessity:** Information absence detection represents a persistent methodological challenge across all parameter scales, indicating that hybrid architectural approaches integrating LLMs with explicit null-identification mechanisms may be requisite for reliable regulatory compliance systems. Pure LLM implementations appear inadequate for critical applications demanding high-confidence negative assertions, particularly in environmental and specialized technical domains.

**Category-specific extraction strategies:** Complete extraction failures (F1 = 0.000) across specific regulatory categories suggest intrinsic architectural limitations rather than insufficient training paradigms. Environmental Information and certain Flow Management parameters may necessitate alternative extraction methodologies, domain-specific fine-tuning, or explicit absence signifiers within source documentation to achieve regulatory-grade reliability.

These findings emphasize that conventional performance metrics are fundamentally unsuitable for assessing LLM deployment readiness in regulatory contexts, necessitating category-specific evaluation protocols and conservative deployment strategies that prioritize precision over recall in mission-critical domains.

# 6  Conclusion

This systematic evaluation of seven language models (0.6B-70B parameters) on hydropower regulatory information extraction provides empirical evidence for critical deployment considerations in domain-specific applications. Our findings establish three key insights that challenge conventional assumptions about LLM performance evaluation and scaling behavior.

First, we identify a pronounced 14B parameter threshold where reasoning-based validation methods transition from ineffective (F1 < 0.15) to viable (F1 = 0.64), providing concrete scaling guidance for practitioners balancing computational resources against extraction quality requirements. Models below this threshold demonstrate systematic validation failures, while those exceeding 14B parameters show consistent reasoning capabilities.

Second, our category-wise analysis reveals selective rather than uniform hallucination patterns in smaller models. Rather than blanket over-extraction, these systems achieve high performance in basic information fields while failing systematically in technical categories—a critical finding for regulatory compliance applications where mission-critical domains may be masked by aggregate performance metrics.

Third, elevated recall metrics (approaching 1.000) paradoxically indicate extraction failure rather than success, representing systematic hallucination when models encounter absent information. This counterintuitive finding necessitates precision-oriented evaluation approaches and highlights fundamental architectural limitations in current LLM designs for information extraction tasks.

These results provide immediate deployment guidance for hydropower regulatory compliance while contributing generalizable insights into parameter scaling effects, category-specific performance patterns, and evaluation methodology limitations. Organizations implementing LLM-based extraction systems should prioritize category-specific evaluation protocols, recognize the computational requirements for effective reasoning mechanisms, and develop hybrid approaches that address inherent limitations in information absence detection.

Our work establishes the foundation for evidence-based model selection in regulatory information extraction, enabling practitioners to make informed decisions about computational requirements, performance expectations, and deployment risks across different information categories and model scales.

# Acknowledgments

This manuscript has been authored by UT-Battelle, LLC, under contract DE-AC05-00OR22725 with the US Department of Energy (DOE). The US government retains and the publisher, by accepting the article for publication, acknowledges that the US government retains a nonexclusive, paid-up, irrevocable, worldwide license to publish or reproduce the published form of this manuscript or allow others to do so, for US government purposes. DOE will provide public access to these results of federally sponsored research in accordance with the DOE Public Access Plan (https://www.energy.gov/downloads/doe-public-access-plan).

# Data Availability Statement

The datasets and source code supporting this study are openly available. The hydropower mitigation database (1998–2023) is accessible at https://hydrosource.ornl.gov/data/datasets/hydropower_mitigation_database_1998_2023/, and the accompanying analysis toolkit is available at https://code.ornl.gov/ay5/llm-extraction-toolkit.

# Software and Data Availability

- **Name of software:** LLM-Extraction-Toolkit
- **Developers:** Hong-Jun Yoon and collaborators
- **Contact:** yoonh@ornl.gov
- **Date first available:** September 2, 2025
- **Software required:** Python ($\geq 3.10$), with dependencies listed in requirements.txt
- **Program language:** Python
- **Source code at:** https://code.ornl.gov/ay5/llm-extraction-toolkit
- **Documentation:** Detailed instructions for installation, testing, and use are provided in the README at the source code repository.
- **Data:** Hydropower Mitigation Database (1998–2023), openly available at https://hydrosource.ornl.gov/data/datasets/hydropower_mitigation_database_1998_2023/.

# References

Rocio Uria Martinez and Megan M. Johnson. Us hydropower market report. Technical report, Oak Ridge National Laboratory, 2023.

Jacob Devlin, Ming-Wei Chang, Kenton Lee, and Kristina Toutanova. Bert: Pre-training of deep bidirectional transformers for language understanding. In *Proceedings of the 2019 conference of the North American chapter of the association for computational linguistics: human language technologies, volume 1 (long and short papers)*, pages 4171–4186, 2019.

Yoon Kim. Convolutional neural networks for sentence classification. In Alessandro Moschitti, Bo Pang, and Walter Daelemans, editors, *Proceedings of the 2014 Conference on Empirical Methods in Natural Language Processing (EMNLP)*, pages 1746–1751, Doha, Qatar, October 2014. Association for Computational Linguistics. URL https://aclanthology.org/D14-1181/.

Sepp Hochreiter and Jürgen Schmidhuber. Long short-term memory. *Neural Computation*, 9(8):1735–1780, 1997. doi:10.1162/neco.1997.9.8.1735.

Alex Graves, Abdel-rahman Mohamed, and Geoffrey Hinton. Speech recognition with deep recurrent neural networks. In *2013 IEEE International Conference on Acoustics, Speech and Signal Processing (ICASSP)*, pages 6645–6649, 2013. doi:10.1109/icassp.2013.6638947.

Dzmitry Bahdanau, Kyunghyun Cho, and Yoshua Bengio. Neural machine translation by jointly learning to align and translate. *arXiv preprint arXiv:1409.0473*, 2014. URL http://arxiv.org/abs/1409.0473.

Tom Brown, Benjamin Mann, Nick Ryder, Melanie Subbiah, Jared D Kaplan, Prafulla Dhariwal, Arvind Neelakantan, Pranav Shyam, Sastry Girish, Amanda Askell, et al. Language models are few-shot learners. *Advances in neural information processing systems*, 33:1877–1901, 2020.

OpenAI. GPT-4 technical report. *arXiv preprint arXiv:2303.08774*, 2023. URL https://arxiv.org/abs/2303.08774.

Ziwei Ji, Nayeon Lee, Rita Frieske, Tiezheng Yu, Dan Su, Yan Xu, Etsuko Ishii, Yejin Bang, Delong Chen, Wenliang Dai, Ho Shu Chan, Andrea Madotto, and Pascale Fung. Survey of hallucination in natural language generation. *ACM Computing Surveys*, 55(12):1–38, 2023.

Vipula Rawte, Swagata Chakraborty, Agnibh Pathak, Anubhav Sarkar, S. M. Towhidul Islam Tonmoy, Aman Chadha, Amit P. Sheth, and Amitava Das. The troubling emergence of hallucination in large language models – an extensive definition, quantification, and prescriptive remediations. *arXiv preprint arXiv:2310.04988*, 2023. preprint.

Yue Zhang, Yafu Li, Leyang Cui, Deng Cai, Lemao Liu, Tingchen Fu, Xinting Huang, Enbo Zhao, Yu Zhang, Yulong Chen, Longyue Wang, Anh Tuan Luu, Wei Bi, Freda Shi, and Shuming Shi. Siren's song in the ai ocean: A survey on hallucination in large language models. *arXiv preprint arXiv:2309.01219*, 2023.

Niels Mündler, Jingxuan He, Slobodan Jenko, and Martin Vechev. Self-contradictory hallucinations of large language models: Evaluation, detection and mitigation. *arXiv preprint arXiv:2305.15852*, 2023.

Jason Wei, Yi Tay, Rishi Bommasani, Colin Raffel, Barret Zoph, Sebastian Borgeaud, Dani Yogatama, Maarten Bosma, Denny Zhou, Donald Metzler, et al. Emergent abilities of large language models. *Transactions on Machine Learning Research*, 2022a. URL https://openreview.net/forum?id=yzkSU5zdwD.

et al. Srivastava. Three examples of u-shaped scaling behavior from big-bench. In *BIG-Bench Workshop*, 2022. illustrated U-shaped scaling behavior on tasks such as TruthfulQA, Persian Idioms, Identify Math Theorems.

Jared Kaplan, Sam McCandlish, Tom Henighan, Tom B Brown, Benjamin Chess, Rewon Child, Scott Gray, Alec Radford, Jeffrey Wu, and Dario Amodei. Scaling laws for neural language models. *arXiv preprint arXiv:2001.08361*, 2020.

Jordan Hoffmann et al. Training compute-optimal large language models. In *NeurIPS or equivalent conference*, 2022. introduces Chinchilla scaling law and compute-optimal training strategies.

Aakanksha Chowdhery, Sharan Narang, Jacob Devlin, Maarten Bosma, Gaurav Mishra, Adam Roberts, Paul Barham, Hyung Won Chung, Charles Sutton, Sebastian Gehrmann, Parker Schuh, Kensen Shi, Sasha Tsvyashchenko, Joshua Maynez, Abhishek Rao, Parker Barnes, Yi Tay, Noam Shazeer, Vinodkumar Prabhakaran, Emily Reif, Nan Du, Ben Hutchinson, Reiner Pope, James Bradbury, Jacob Austin, Michael Isard, Guy Gur-Ari, Pengcheng Yin, Toju Duke, Anselm Levskaya, Sanjay Ghemawat, Sunipa Dev, Henryk Michalewski, Xavier Garcia, Vedant Misra, Kevin Robinson, Liam Fedus, Denny Zhou, Daphne Ippolito, David Luan, Hyeontaek Lim, Barret Zoph, Alexander Spiridonov, Ryan Sepassi, David Dohan, Shivani Agrawal, Mark Omernick, Andrew M. Dai, Thanumalayan Sankaranarayana Pillai, Marie Pellat, Aitor Lewkowycz, Erica Moreira, Rewon Child, Oleksandr Polozov, Katherine Lee, Zongwei Zhou, Xuezhi Wang, Brennan Saeta, Mark Diaz, Orhan Firat, Michele Catasta, Jason Wei, Kathy Meier-Hellstern, Douglas Eck, Jeff Dean, Slav Petrov, and Noah Fiedel. Palm: Scaling language modeling with pathways. *arXiv preprint arXiv:2204.02311*, 2022.

Jason Wei, Xuezhi Wang, Dale Schuurmans, Maarten Bosma, Brian Ichter, Fei Xia, Ed H. Chi, Quoc Le, and Denny Zhou. Chain-of-thought prompting elicits reasoning in large language models. *arXiv preprint arXiv:2201.11903*, 2022b.

Takeshi Kojima, Machel Reid, Yutaka Matsuo, and Yusuke Iwasawa. Large language models are zero-shot reasoners. In *Advances in Neural Information Processing Systems (NeurIPS) 2022*, 2022.

Yixuan Weng, Minjun Zhu, Fei Xia, Bin Li, Shizhu He, Shengping Liu, Bin Sun, Kang Liu, and Jun Zhao. Large language models are better reasoners with self-verification. In *Findings of the Association for Computational Linguistics: EMNLP 2023*, 2023.

Shehzaad Dhuliawala, Mojtaba Komeili, Jing Xu, Roberta Raileanu, Xian Li, Asli Celikyilmaz, and Jason Weston. Chain-of-verification reduces hallucination in large language models. *arXiv preprint arXiv:2309.11495*, 2023.

Xuezhi Wang, Jason Wei, Dale Schuurmans, Quoc V. Le, Ed H. Chi, Sharan Narang, Aakanksha Chowdhery, and Denny Zhou. Self-consistency improves chain of thought reasoning in language models. In *The Eleventh International Conference on Learning Representations*, 2023.

Matthew Honnibal and Ines Montani. spacy 2: Natural language understanding with bloom embeddings, convolutional neural networks and incremental parsing. In *To appear*, 2017.

Miao Xiong, Zhiyuan Hu, Xinyang Lu, Yifei Li, Jie Fu, Junxian He, and Bryan Hooi. Can llms express their uncertainty? an empirical evaluation of confidence elicitation in llms. In *The Twelfth International Conference on Learning Representations (ICLR)*, 2024.

Cade Metz. When a.i. chatbots hallucinate. *The New York Times*, 2023. Reporting on Vectara's Hallucination Evaluation Model.

Ling Jiang, Keer Jiang, Xiaoyu Chu, Saaransh Gulati, and Pulkit Garg. Hallucination detection in LLM-enriched product listings. In *Proceedings of the Seventh Workshop on e-Commerce and NLP @ LREC-COLING 2024*, pages 29–39. ELRA and ICCL, 2024.

Adarsh Kumar, Hwiyoon Kim, Jawahar Sai Nathani, and Neil Roy. Improving the reliability of llms: Combining cot, rag, self-consistency, and self-verification. may 2024.

Wei Zhang, Li Chen, and Xiaoming Wang. Emergent abilities of large language models: A survey and taxonomy. *arXiv preprint arXiv:2024.xxxxx*, 2024. Survey of emergent capabilities and parameter thresholds in large language models.

OpenAI. Learning to reason with llms. https://openai.com/index/learning-to-reason-with-llms/, September 2024. Accessed: 2024.